\documentclass[11pt,a4paper]{article}
\usepackage{emnlp-ijcnlp-2019}
\usepackage{times}
\usepackage{latexsym}
\usepackage{amsmath}
\usepackage{amssymb}
\usepackage{graphicx}
\usepackage{algorithm}
\usepackage[noend]{algpseudocode}
\usepackage{setspace}
\usepackage{url}
\usepackage{subcaption}
\usepackage{tabularx}
\usepackage{xcolor}
\definecolor{darkgreen}{HTML}{1D80DB}

\usepackage{booktabs}
\usepackage{statistics-macros}

\usepackage[normalem]{ulem}

\aclfinalcopy 

\definecolor{lightbl}{RGB}{67,162,202}

\newcommand{\ptrain}{p_{\text{z}}^{\text{train}}}
\newcommand{\padv}{p_{\text{z}}}
\newcommand{\ptest}{p_{\text{z}}^{\text{test}}}
\newcommand{\ptrainx}{p_{\text{x}}^{\text{train}}}
\newcommand{\ptestx}{p_{\text{x}}^{\text{test}}}
\newcommand{\padvx}{p_{\text{x}}}
\newcommand{\potherx}{p_{\text{x}}^{\text{other}}}
\newcommand{\phattrain}{\hat{p}_{\text{z}}^{\text{train}}}

\newcommand{\pwdist}[1]{p_{\text{x}\mid #1}}
\newcommand{\pwprob}[2]{p_{\text{x}\mid{#2}}\left(#1\;\middle|\;#2\right)}
\newcommand{\pset}{\mathcal{P}_{\text{z}}^{\alpha}}
\newcommand{\pmodel}{p_\theta}
\newcommand{\pbase}{p_\beta}
\newcommand{\iter}[1]{{^{(#1)}}}
\newcommand{\psetx}{\mathcal{P}_{\text{x}}^{\alpha}}

\newcommand{\atr}{\alpha^*}       

\usepackage{etoolbox}

\title{Distributionally Robust Language Modeling}

\author{Yonatan Oren*$^{1}$ ~~ Shiori Sagawa*$^{1}$ ~~ Tatsunori B. Hashimoto*$^{1,2}$ ~~ Percy Liang$^{1}$ \\
 \textbf{(* equal contribution)} \\
 $^1$Stanford Computer Science ~~ $^2$Stanford Statistics\\
 {\small \tt \{yonatano,thashim\}@stanford.edu  ~~ \{ssagawa,pliang\}@cs.stanford.edu} \\
}

\date{}

\begin{document}
\maketitle

\begin{abstract}
Language models are generally trained on data spanning a wide range of topics (e.g., news, reviews, fiction),
but they might be applied to an a priori unknown target distribution (e.g., restaurant reviews).
In this paper, we first show that training on text outside the test distribution can degrade test performance when using standard maximum likelihood (MLE) training.
To remedy this without the knowledge of the test distribution, we propose an approach which trains a model that performs well over a wide range of potential test distributions. 
In particular, we derive a new distributionally robust optimization (DRO) procedure which minimizes the loss of the model over the \emph{worst-case} mixture of topics with sufficient overlap with the training distribution. 
Our approach, called \emph{topic conditional value at risk (topic CVaR)}, obtains a 5.5 point perplexity reduction over MLE when the language models are trained on a mixture of Yelp reviews and news and tested only on reviews.

\end{abstract}

\section{Introduction} 
\label{intro}

Large-scale language modeling plays a central role in both text generation
\cite{sordoni2015neural,nallapati2016abstractive}
and unsupervised pre-training 
\cite{vaswani2013decoding, dai2015semi, mccann2017learned, peters2018elmo,devlin2018BERT, radford2018improving}. In both settings, a single language model is trained on a large corpus containing a range of topics (e.g. news, fiction, and reviews). This language model is then applied in many different tasks, each with a specific test distribution (e.g., analyzing the sentiment of restaurant reviews). 
Can we train a single general-purpose language model that works across a wide range of potential test distributions?

\begin{figure}
\centering
\includegraphics[scale=0.2]{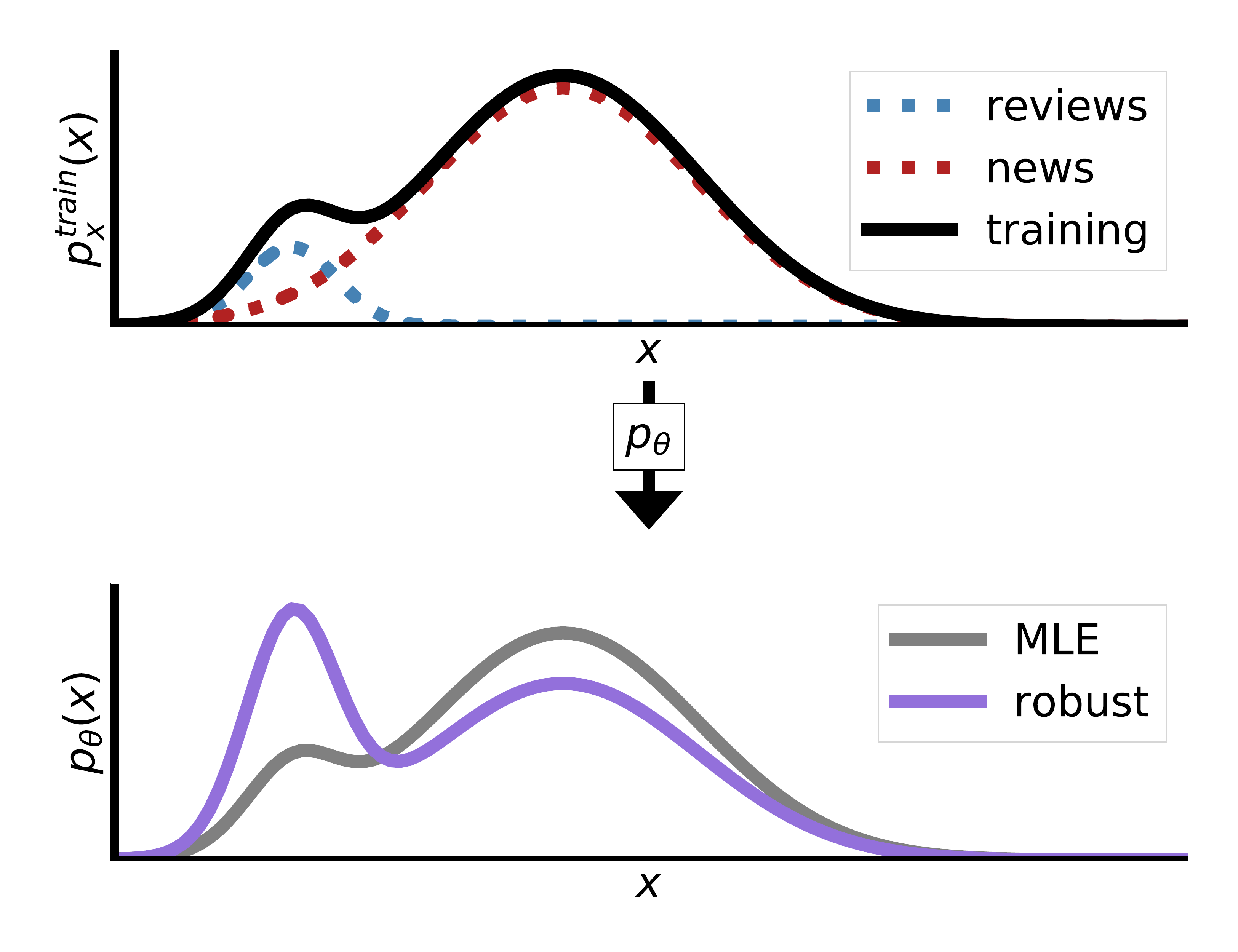}
\caption{
  Illustration of a training corpus as a density (black) with mostly news stories (red) and a small number of restaurant reviews (blue).
  The standard MLE model (gray) reflects the underlying data and assigns little weight to reviews,
  and thus performs poorly on reviews.
  A more robust model should try to equalize the weight across all topics so that it can perform well regardless of which topics appear at test time.
  }
  \label{fig:figone}
  \vspace{-10pt}
\end{figure}

In this work, we first demonstrate that standard maximum likelihood training on a large, heterogeneous dataset can fail to achieve this goal.
While more data is generally better, the presence of text outside the target distribution actually \emph{degrades} performance on a target test distribution.
For example, a language model trained on Yelp reviews achieves a perplexity of 32,
and this perplexity increases to 43 when trained on a mixture of 10\% Yelp and 90\% newswire sentences from the One Billion Word Benchmark \cite{chelba2013one}.
Performance degrades because
existing maximum likelihood estimation (MLE) objectives tend to emphasize model performance on more common sentences and topics at the expense of infrequent ones (Figure \ref{fig:figone}). 

While the above performance degradation can be mitigated by fine-tuning and domain adaptation techniques
\cite{shimodaira2000improving,quinonero2009dataset,daume07easyadapt,ben2010theory,blitzer2011domain, pryzant2017domainmix,ganin2015domain,tzeng2014domain},
these methods require knowing the test distribution and training a separate model specific to each target distribution.
Instead, we aim to train a \emph{single} model that performs well across many unknown test distributions.

In order to do this, we will train a model that performs uniformly well over an entire family of potential test distributions.
Since we cannot expect to do well on all possible test distributions,
we consider the \emph{subpopulation shift} setting, in which the test distribution is a subpopulation of the training distribution, 
and seek good performance across all such test distributions
(e.g. Yelp reviews in a Yelp-newswire mixture). \footnote{The subpopulation assumption refers to overlaps in \emph{distributions}, rather than individual examples. Our assumptions do not require overlap in the training and test data.} In other words, adding data from topics outside the test topics should not hurt. It seems reasonable to protect against subpopulation shifts, 
intuitively because large-scale data collection schemes are designed to cover a diverse array of topics as a way to generalize to potential test distributions.

We train a model that performs well over all subpopulations by minimizing the risk for the \emph{worst-case} subpopulation, following the distributionally robust optimization (DRO) literature \cite{bental2013robust}.
While an existing DRO framework called the conditional value at risk (CVaR)
ensures uniformly good performance across subpopulations \cite{rockafellar2000optimization,duchi2018learning}, 
we demonstrate that na\"ively applying this approach to language modeling fails due to three challenges.
First, the existing CVaR approach is too conservative because it considers robustness to \emph{arbitrary} subpopulations.
Such worst-case subpopulations are attained by adversarially choosing the hardest, most unusual sentences.
Instead, we propose to consider \emph{meaningful} subpopulations, defined by topics in a corpus \cite{hu2018does}.
Second, applying CVaR directly to log loss results in a loss which is biased towards topics with high entropy, instead of those for which the model performs poorly relative to what is possible. We correct this by introducing a new \emph{baselined} loss function which measures losses relative to the entropy of each topic. 
Finally, existing optimization algorithms for CVaR are either inapplicable to topic-based robustness sets or unscalable because they require batch optimization.
We develop a scalable online algorithm which identifies the worst-performing topics at each iteration and upweights examples from those topics.

With these methodological improvements, we demonstrate that our approach, \emph{topic CVaR}, improves robustness against subpopulation shifts.
Topic CVaR reduces perplexity on the Yelp review corpus by 5.5 points compared to MLE when trained on the Yelp-One Billion Word Benchmark mixture from before.
We also show improved robustness even when the shift is not strictly a subpopulation shift.
Topic CVaR also achieves a 4 point perplexity reduction on a test distribution (TripAdvisor hotel reviews) that is similar to, but not strictly a subpopulation of the training distribution (Yelp and newswire text).

\section{Problem Statement} 
\label{problemst}
Our goal is to learn a language model $\pmodel$ based on sentences sampled from the training distribution $x \sim \ptrainx$, such that $\pmodel$ performs well on unknown test distributions $\ptestx$.

Language models $\pmodel$ are generally trained to approximate $\ptrainx$ by minimizing the KL divergence $\dkl{\ptrainx}{\pmodel}$ via maximum likelihood estimation (MLE),
\begin{align}
  \label{opt:klmle}
 \inf_\theta \E\left[- \log\pmodel(x)\right].
\end{align}

When $\ptestx = \ptrainx$, classical statistical theory guarantees that a model trained via MLE performs well on the test distribution given sufficient data.
However, when $\ptestx$ is not identical to $\ptrainx$, MLE can perform poorly no matter how much data is observed.
This is because the test set might consist solely of sentences from topics that are infrequent during training, to which MLE would assign low probabilities.

\begin{figure}
\centering
\includegraphics[scale=0.25]{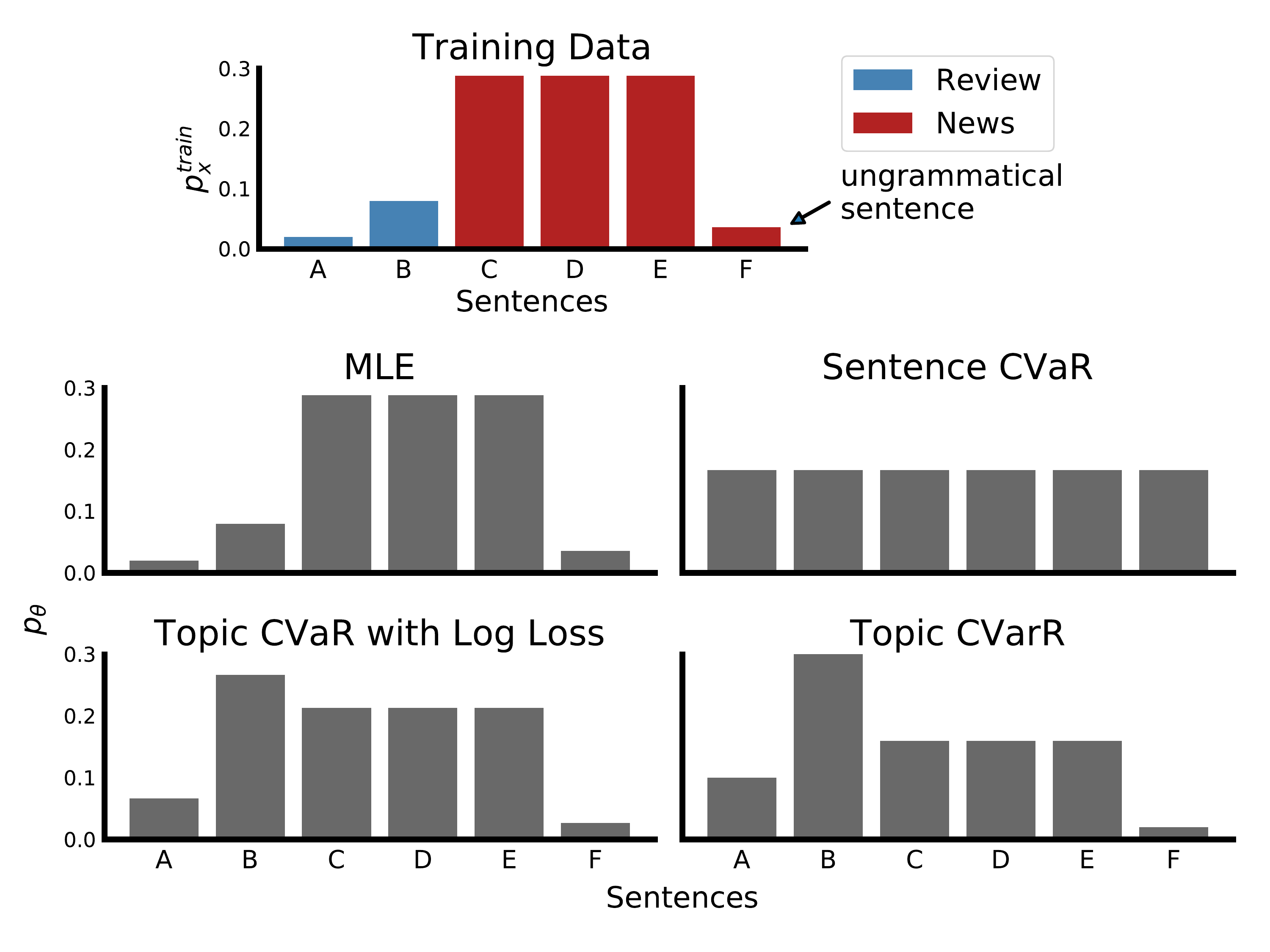}
\caption{
  Toy example of a multinomial distribution over six sentences (top). Different panels illustrate models learned by different training procedures.
    MLE fits common topics (news) at the expense of rare ones (reviews).
    Sentence CVaR is too conservative, overemphasizing the ungrammatical sentence.
    Topic CVaR with log loss overemphasizes difficult topics (news) over easy ones (review).
    Topic CVaR (with baselining) balances the weight assigned to each topics, as desired.
  }
  \vspace{-7pt}
\label{fig:example}
\end{figure}

To illustrate this point, consider the toy example drawn in Figure \ref{fig:example}.
In this example, the training distribution $\ptrainx$ is a multinomial distribution over six possible sentences A--F, with two from reviews and four from news. Sentence F is ungrammatical and thus has an extremely low probability.
The training distribution includes $10\%$ reviews and $90\%$ news, whereas the test distribution could be all reviews, all news, or a mixture.
MLE assigns low probabilities to any review and thus performs poorly when evaluated solely on reviews.
To be robust, we intuitively need a more conservative objective that encourages models to assign higher probabilities to rare but valid sentences.

In order to achieve this, we want to learn a model $\pmodel$ which performs well in situations where $\ptrainx \neq \ptestx$ for a large set of potential test distributions $\mathcal{P}$, termed the uncertainty set.
By training a model that performs well on all distributions in the uncertainty set $\mathcal{P}$, we can ensure good test performance as long as $\ptestx \in \mathcal{P}$. 

More formally, this approach falls under the framework of distributionally robust optimization (DRO) \cite{bental2013robust}.
With DRO, we optimize a model for loss $\ell$ and a set of potential test distributions $\mathcal{P}$ by minimizing the risk under the \emph{worst-case} distribution in $\mathcal{P}$,
\begin{align}
    \sup_{\padvx \in\mathcal{P}} \E_{\padvx}[\ell(x;\theta )].
\end{align}
Observe that the above worst-case objective does not depend on the unknown quantity $\ptestx$.
The objective also upper bounds the test risk for all $\ptestx \in \mathcal{P}$ as
\begin{align}
\label{klgoal}
    \E_{\ptestx}[\ell(x;\theta)] \leq \sup_{ \padvx \in\mathcal{P}} \E_{\padvx}[\ell(x;\theta )],
\end{align}
so optimizing the above objective gives guarantees on test performance whenever $\ptestx \in \mathcal{P}$.

DRO provides a conceptually appealing framework for learning under train-test mismatch. However, it crucially depends on both the choice of uncertainty set $\mathcal{P}$ and loss $\ell$, and we will discuss these choices in the next section.

\section{Robust Language Modeling} 
\label{approach}

We will begin by applying standard distributionally robust optimization approaches to the log loss (Section \ref{approach:joint}), and showing that this na\"ive approach suffers from two drawbacks:
\begin{enumerate}
  \itemsep=0pt
  \item Existing DRO uncertainty sets $\mathcal{P}$ are too conservative.
  \item The log loss overemphasizes topics with inherently high entropy.
\end{enumerate}
These drawbacks will motivate our development of a new approach we call topic CVaR, which addresses these two problems (Sections \ref{approach:topics} and \ref{approach:loss}).

\subsection{Robustness to arbitrary subpopulations}
\label{approach:joint}
Observing that MLE is not robust because it assigns low probabilities (i.e. incurs high losses) on rare sentences, we might initially try to define $\mathcal{P}$ as individual training examples to ensure low loss on \emph{all} data points.  However, this is far too conservative, since the worst-case distribution would consist of exactly one data point. Therefore, we may want to optimize a slightly more realistic uncertainty set consisting of all sufficiently large subpopulations of the training distribution.

Minimizing losses over all subpopulations of the training distribution can be formulated as a type of distributionally robust optimization (DRO) problem \cite{duchi2018learning}, which has been used to regularize models \cite{duchi2016variance}, defend against adversarial examples \cite{sinha2018certifiable}, and improve the fairness of models \cite{hashimoto2018repeated}.

One type of distributionally robust loss is known as conditional value at risk (CVaR) which guarantees low losses on all $\alpha$-fraction subpopulations of the training distribution \cite{rockafellar2000optimization}. This corresponds to defining the uncertainty set $\mathcal{P}$ as all sentence distributions that are \emph{$\alpha$-covered} by $\ptrainx$,
\begin{align}
    \label{def:psetx}
    \psetx \defeq \{\padvx : \alpha\padvx(x) \le \ptrainx(x) ~~ \forall x\}.
\end{align}
This is equivalent to defining $\psetx$ as the set of $\padvx$ which fulfills $\ptrainx = \alpha\padvx + (1-\alpha)\potherx$ for some distribution $\potherx$.

To achieve low loss on all possible test distributions in $\psetx$, we minimize the expected loss under the worst-case distribution, 
\begin{align}
\label{opt:joint}
    \sup_{\padvx\in\psetx} \E_{x\sim\padvx}[\ell(x;\theta)].
\end{align}
For the remainder of the paper, we will refer to this approach as \emph{sentence CVaR}, highlighting the fact that it considers robustness over arbitrary sets of sentences.
It intuitively encourages uniform performance across all subpopulations of sentences by downweighting sentences with low loss, and upweighting sentences with high loss.

Because sentence CVaR considers \emph{arbitrary} groups of examples, it can be too conservative in our problem setting. 
While sentence CVaR can prevent modeling common sentences at the cost of rare ones, it can also encourage modeling invalid sentences at the expense of valid ones.
Returning to our example in Figure \ref{fig:example} with $\ell(x;\theta)=-\log \pmodel(x)$ , sentence CVaR with for sufficiently low $\alpha$ achieves perfectly uniform performance. It equalizes likelihoods across all sentences, which unfortunately also results in high probabilities assigned to the ungrammatical sentence F.

\subsection{Robustness over Topics}
\label{approach:topics}
Sentence CVaR is too conservative since it allows for arbitrary groups --- including ones consisting of purely invalid sentences. To remedy this, we will optimize models for all \emph{meaningful} subpopulations instead of \emph{arbitrary} ones.

One way to achieve this is through robustness over topics, rather than individual examples.
For example, a news corpus often contains a variety of topics (politics, business, opinion, food) and a test corpus may contain these topics with different proportions.
A robust language model should perform well on a wide range of topic mixtures without taking the topic identity as an input.

Formally, 
we posit that each sentence $x$ belongs to some latent topic $z$, which has a sentence distribution $\pwdist{z}$.
We want our models to be robust to shifts in the topic distribution, where we have $z\sim\ptrain$ and $z\sim\ptest$.
In this case, we can define a natural uncertainty set for CVaR, defined over latent topics rather than individual examples.
Extending the definition of $\alpha$-covered distributions to topics, we have the set 
\begin{align}
\label{def:pset}
\pset \defeq \{\padv : \alpha\padv(z)\le\ptrain(z) ~~\forall z\}
\end{align}
and the objective is the expected loss under the worst-case topic distribution,
\begin{align}
\label{opt:marg}
\sup_{\padv\in\pset} \E_{z\sim\padv}\left[\E_{x\sim\pwdist{z}}\left[\ell(x;\theta)\right]\right].
\end{align}
This objective intuitively encourages uniform loss across topics by upweighting topics incurring high losses and downweighting topics with low losses, while keeping the conditional distribution of sentences given a topic constant.

\subsection{Baselined Loss Function}
\label{approach:loss}

Recall that DRO depends critically on the choice of uncertainty set and loss function. Having specified the uncertainty set, we now turn to the choice of loss $\ell(x;\theta)$. While the log loss $\ell(x;\theta) = -\log\pmodel(x)$ is the standard choice in language modeling, we show that this approach has a flaw in the robust setting and propose a corrected loss.

\paragraph{Log Loss.}
Using log loss on CVaR encourages uniform \emph{absolute} log-likelihoods across topics even if some topics are much harder than others. 
For example, consider a model which performs nearly optimally on difficult topics and highly suboptimally on easy topics. Since log loss measures absolute performance, it would force the model to focus on the difficult topic \emph{even if the model can't improve further on this topic}. 
In the example in Figure \ref{fig:example}, news is emphasized over reviews because news has higher entropy and thus higher difficulty. Empirically, we observe that log loss with CVaR forces the models to
 focus almost entirely on the difficult topics such as long news stories.

 \paragraph{Baselined Loss.}
We now propose a new baselined loss, which encourages uniform \emph{relative} performance across topics.
We refer to our approach with the baselined loss as \emph{topic CVaR}.

The baselined loss function $\ell(x,z;\theta) = \log\pwprob{x}{z}-\log\pmodel(x)$ evaluates the performance of the model relative to the best possible model for the topic, $\log\pwprob{x}{z}$. Although we do not observe $\log\pwprob{x}{z}$, we will show later in section \ref{alg:entropy} that we can estimate sufficient statistics of $\log\pwprob{x}{z}$ that allow us to compute the baselined loss.
By using baselined loss, we intuitively encourage models to perform as well as it can on each topic while making optimal trade-offs among topics.

Plugging the baselined loss into the robust objective \eqref{opt:marg}, the optimization problem is 
\begin{align}
\label{opt:drobase}
    \sup_{\padv\in\pset} \E_{z\sim\padv}\left[\E_{x\sim\pwdist{z}}\left[\log\pwprob{x}{z}-\log\pmodel(x)\right]\right], 
\end{align}
which can be simplified to 
\begin{align}
\label{opt:drokl}
    \sup_{\padv\in\pset} \E_{z\sim\padv}\left[\dkl{\pwdist{z}}{\pmodel}\right].
\end{align}
Topic CVaR thus minimizes the per-topic KL divergences, and
this interpretation fits nicely with a general goal of training $\pmodel$ that matches the test distribution.
Unlike in the MLE case, minimizing the KL is not equivalent to minimizing the log loss.
In MLE, minimizing $\text{KL}(\ptrainx\|\pmodel)=\E\left[\log\ptrainx(x)-\log\pmodel(x)\right]$ is equivalent to minimizing the log loss because $\log\ptrainx(x)$ can be treated as a constant.
However, in topic CVaR, the analogous baseline entropy term $\log\pwprob{x}{z}$ depends on $z$ and thus is not a constant with respect to the outer supremum.

In the running toy example (Figure \ref{fig:example}), topic CVaR results in robust models that perform relatively well on both news and reviews.
The resulting model is a mixture of news and review distribution with equal weights on the two topics. 

In summary, topic CVaR contains two improvements over existing DRO approaches: using the latent topic distribution $\ptrain$ to specify the uncertainty set and defining the baselined loss. In the following section, we will describe an algorithm which optimizes this topic CVaR objective.

\section{Algorithm}
\label{algorithm}
We now operationalize the principles in the previous section, specifying (i) how we choose topics (Section~\ref{alg:topics}), (ii) how we estimate the baseline (Section~\ref{alg:entropy}), and (iii) how to efficiently optimize the robust objective \eqref{opt:marg} (Section~\ref{alg:opt}).

\subsection{Identifying Topics}
\label{alg:topics}
The topic CVaR objective requires topic assignments $z$ for each sentence in order to define the uncertainty set $\mathcal{P}$. Since the topics determine the set of $\ptestx$ distribution for which the model performs well, 
we seek topics whose subpopulation shifts capture realistic potential test settings. 

We use latent Dirichlet allocation (LDA) \citep{blei03lda} to cluster the sentences into latent topics. LDA assigns each word in a sentence to a topic, and we assign each sentence to the topic with highest total posterior probability.

\subsection{Estimating Baselined Losses}
\label{alg:entropy}
Recall that topic CVaR uses KL-divergence as the loss term (Eq.~\eqref{opt:drokl}),
\begin{multline*}
  \dkl{\pwdist{z}}{\pmodel} := \E_{\pwdist{z}}[\log\pwprob{x}{z}] \\
  - \E_{\pwdist{z}}[\log\pmodel(x)].
  \end{multline*}
While we can estimate the log loss term $ \E[\log\pmodel(x)]$ from samples, the entropy term $H(X \mid Z=z) := \E_{\pwdist{z}}[-\log\pwprob{x}{z}]$ is not something we can easily estimate.

We thus propose to estimate the entropies $H(X\mid Z=z)$ by fitting a \emph{baseline model} $\pbase$ for each topic, and computing $H_{\beta}(X\mid Z=z) := \E_{\pwdist{z}}[-\log\pbase(x\mid z)]$.\footnote{$H_{\beta}$ yields accurate solutions to the topic CVaR problem as long as they capture the entropy up to a constant (i.e. $H_\beta(X \mid Z = z) \approx H(X \mid Z = z) + c$)}
In practice, we use a bigram model, which was fast enough to scale and worked sufficiently well in experiments.


\subsection{Online Optimization of topic CVaR}
\label{alg:opt}
No scalable, online algorithm exists for optimizing the topic CVaR objective.
Many DRO problems admit efficient batch optimization procedures based on Lagrangian duality \cite{duchi2016}.
However, this approach fails for topic CVaR, since the dual form requires exact computations rather than stochastic estimates of $\E_{\pwdist{z}}\left[- \log\pmodel(x)\right]$. Online algorithms for DRO exist \citep{namkoong2016stochastic}, but do not handle the nested maximization-expectation structure arising in topic CVaR (Eq.~\eqref{opt:marg}).

Because of this, we develop an online optimization procedure for topic CVaR compatible with stochastic gradient descent methods. 
The topic CVaR problem is a two-player minimax game between the model parameter $\theta$ and the potential test distribution $\padv$.
Intuitively, $\padv$ attempts to be the worst-case distribution and maximize the robust objective, while $\theta$ attempts to minimize the robust objective.
The precise two-player minimax game is 
\begin{align}
\label{opt:game}
\inf_\theta \sup_{\padv \in \pset} \E_{z\sim\padv}\left[L(z;\theta)\right],
\end{align}
where the expected loss for each $z$ (inner expectation) is $L(z;\theta):=\E_{x\sim\pwdist{z}}\left[\ell(x;\theta)\right]$.

In the above two-player game, the game proceeds in multiple rounds $t=1,2,\dots$.
At each round, the players select $\padv\iter{t}$ and $\theta\iter{t}$.
It is standard to interleave parameter updates between the two players in minimax optimization, and we describe the precise update rules in subsequent paragraphs.
To carry out these updates, we keep track of an empirical estimate of the probability $\ptrain(z)$ at each iteration $t$, which we refer to as $\phattrain\iter{t}(z)$.
We also keep track of the historical average of losses incurred for each topic so far, up to the current round $t$, which we call $\hat{L}\iter{t}(z;\theta\iter{1:t})$.
Concretely, $\hat{L}\iter{t}(z;\theta\iter{1:t})$ is computed as an average of $\{\ell(x\iter{t^\prime};\theta\iter{t^\prime} ) : t^\prime\in[t], z\iter{t^\prime}=z\}$.

At each iteration $t$, $\padv$ is updated by selecting an optimal value with respect to historical losses up to the current iteration, loosely inspired by the ``Be The Leader" algorithm. This results in the following update rule to $\padv$,
\begin{align}
\label{btl}
\padv\iter{t} = \argmax_{\padv\in\pset} \E_{z\sim\padv}\left[\hat{L}\iter{t}(z;\theta\iter{1:t})\right].
\end{align}
The above $\argmax$ can computed efficiently by ordering topics in the order of decreasing average loss, and assigning each topic either $\frac{\phattrain(z)}{\alpha}$ or the probability left to be assigned, whichever is lower.\footnote{For example with $\alpha=0.2$, $\hat{L}\iter{t} = [40, 30, 60]$, and $\phattrain= [0.2, 0.8, 0.1]$, then $\padv\iter{t}=[0.5, 0, 0.5]$.}

We update $\theta$ with online gradient descent,
\begin{align*}
  \label{ogd}
\theta\iter{t} = \theta\iter{t-1} - \epsilon\frac{\padv\iter{t}(z\iter{t})}{\phattrain\iter{t}(z\iter{t})}\nabla\ell(x\iter{t};\theta\iter{t-1}),
\end{align*}
where $\epsilon$ is the learning rate.

To give intuition for the two updates, first note that $\frac{\padv\iter{t}(z)}{\ptrain\iter{t}(z)}=\frac{1}{\alpha}$ on approximately $\alpha$ fraction of the data and this ratio acts as an indicator function which determines if an example is part of the worst-case set or not. If it is, we update the model and otherwise we ignore it.



\section{Experiments \label{experiments}} 

\newcommand{\lmdata}{\textsc{OneBWord}}

We demonstrate that topic CVaR improves maximum likelihood language models when $\ptrainx\neq\ptestx$. Section \ref{expdetails} outlines the experimental setup while Section \ref{robustsub} shows the robustness improvements and analysis of topic CVaR.

\subsection{Evaluation Details}
\label{expdetails}

\paragraph{Datasets.}
We use the following three corpora:
the Yelp review corpus (\textsc{Yelp}, \shortcite{yelp2017yelp}), 
One Billion Word benchmark corpus (\lmdata), and the TripAdvisor Annotated Dataset (\textsc{TripAdv}, \citet{marcheggiani2014hierarchical}).

We preprocess the corpora using \textsc{SpaCy}  (\citet{honnibal2015nmdp}) by removing sentences with fewer than 10 characters, segmenting sentences, tagging named-entities, and replacing each entity with its corresponding OntoNotes tag.

\paragraph{Vocabulary.}
Our experiments will evaluate models using perplexity, which depends on the choice of vocabulary. To make perplexity comparable for models trained on different datasets, we use a single, fixed vocabulary formed by combining the most frequently occurring $10,000$ words in each corpus.
All words in the mixtures which are not in the vocabulary (1$-$3\% in our experiments) are replaced with a special \textit{unk} token.

\paragraph{Clustering.} To cluster sentences in the training set, we ran LightLDA (\citet{yuan2015lightlda}) for 100 iterations with prior hyperparameters $\alpha=0.1$, $\beta=1.0$ and $2$ Metropolis-Hastings steps. We set the model to find $10$ topics, as this resulted in stable clusters consisting of semantically similar sentences.

\paragraph{Models.}
Our models are Transformer \cite{vaswani2017attention} based language models trained using the \textsc{Fairseq} sequence-to-sequence toolkit \shortcite{gehring2017convolutional}. We use the same model architecture, optimizers, and hyperparameters for both MLE and CVaR. For both models, we use Nesterov's accelerated gradient descent, a fixed learning rate of 0.01, minibatch size of 500 sentences, and 30k minibatches (corresponding to 100 epochs on the \textsc{Yelp} corpus). These values were derived by tuning a MLE model trained on the \textsc{Yelp} data and tested on the \textsc{Yelp} dev set.

\paragraph{Hyperparameters}
Topic CVaR can be unstable at small $\alpha$ values due to the fact that we are optimizing for worst-case errors. Because of this, we make three small but important modifications to the algorithm. (i) We use $\alpha=0.2$ to estimate models for $\atr < 0.2$, as small $\alpha$s can cause gradients to become unstable; (ii) we set a minimum $\padv(z)/\ptrain(z)$ value of 0.1; and (iii) we compute historical losses using exponentially weighted moving averages. With these modifications, the model reliably converges to similar validation losses.

 \subsection{Language Model Robustness}
 \label{robustsub}

 We seek to assess the performance of MLE and CVaR models under subpopulation shift. In order to do this, we train language models on various mixtures of \textsc{Yelp} and \lmdata{} corpora and evaluate the models on a held-out set of \textsc{Yelp} sentences.

 We will construct a training corpus, whose distribution \emph{$\atr$-covers} the test distribution (i.e. $\atr$ fraction of the training distribution corresponds to the Yelp distribution). In this case, we expect that topic CVaR with $\alpha = \atr$ to perform well since the test set exactly fulfills the subpopulation shift assumption.

 To form a training corpus, whose distribution $\atr$-covers the \textsc{Yelp} distribution, we mix a fixed set of 500,000 sentences from \textsc{Yelp} training subset with $500,000(1 - \atr)/\atr$ sentences from \lmdata. This results in a dataset where $\atr$ of the training data comes from \textsc{Yelp}.
 The test corpus is composed of sentences from the \textsc{Yelp} test subset, with no sentence overlap with the training corpora. 
 Since the absolute number of \textsc{Yelp} samples in the training corpora remains constant across different values of $\atr$, we expect that a model which is robust to added nuisance data will perform equally well on a \textsc{Yelp}-only test set, even as the mixture proportion of \lmdata{} samples in the training corpus increases.

 \paragraph{Oracle model.} We estimate the \emph{oracle} performance of a robust language model as running topic CVaR where the topic $z = \{\textsc{Yelp}, \lmdata{}\}$ and the topic assignments use the ground truth corpus identities rather than a clustering algorithm. In this case, when $\atr=\alpha$ we are directly minimizing the worst-case baselined test loss over both \textsc{Yelp} and \lmdata{}.

   \begin{figure}[ht!]
    \centering
  \includegraphics[scale=0.50]{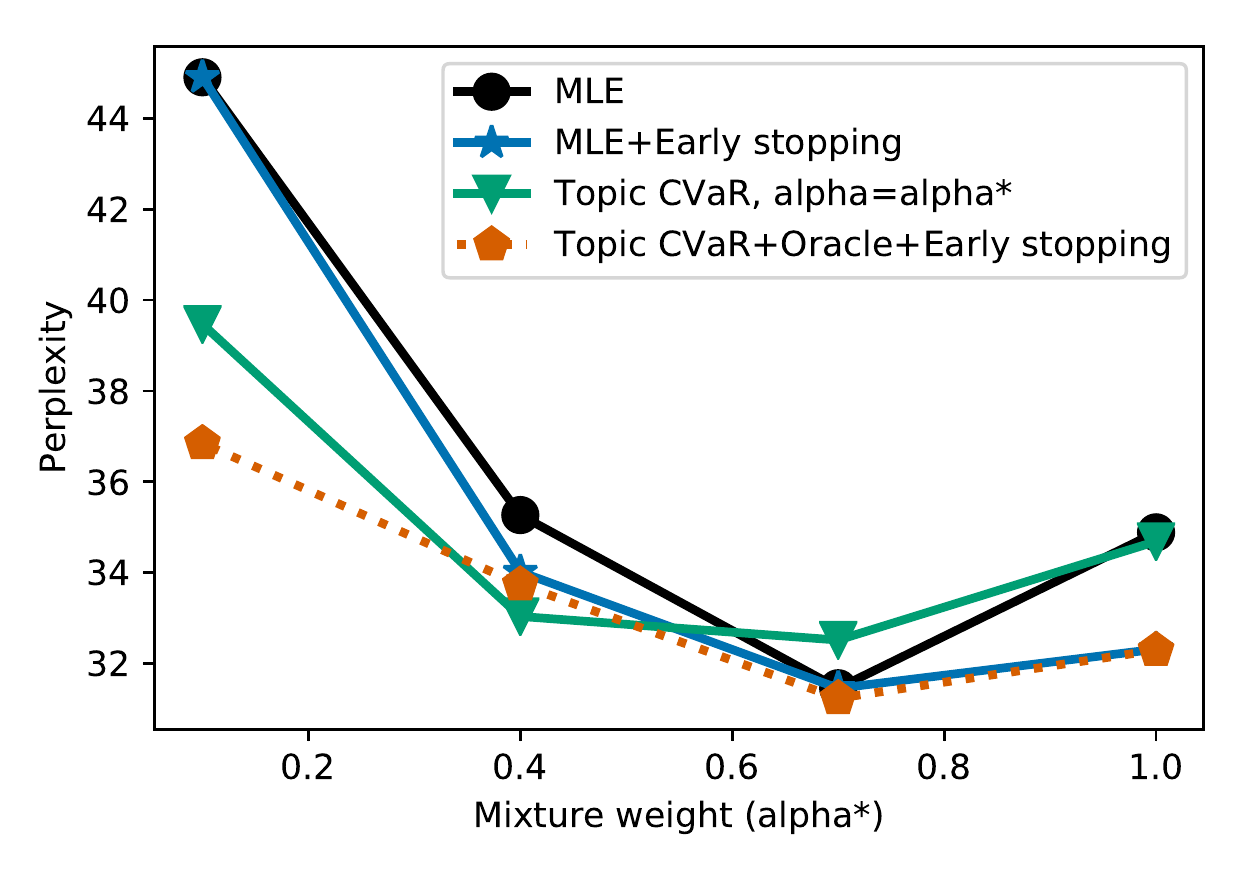}
   \caption{Topic CVaR (green) provides substantial improvements in perplexity compared to MLE (black and blue) as the amount of train-test mismatch increases ($\atr\to 0$). This performance is close to the oracle performance which uses ground truth corpus labels and early stopping (orange). 
   }
  \label{fig:droplots}
\end{figure}

   \paragraph{Topic CVaR improves robustness over MLE.} Using the \textsc{Yelp}-\lmdata{} mixtures, we evaluate the robustness of topic CVaR and MLE to added nuisance data. We find that with no nuisance data, the MLE model matches the topic CVaR model (Figure \ref{fig:droplots} $\atr=1.0$).
  As we add data from $\lmdata{}$ and $\atr$ decreases to 0.7, we find some \emph{positive transfer} effects where the increased data from the \lmdata{} corpus improves the performance on Yelp. However, as the fraction of nuisance data grows further and $\atr$ drops below 0.4 the MLE models suffer large increases in perplexity, incurring up to 10 additional points of perplexity. Early stopping according to validation perplexity on \textsc{Yelp} does not improve this substantially beyond the basic MLE model (blue star). On the other hand, applying topic CVaR with $\atr=\alpha$ provides substantial boosts to language model performance for small $\atr$, with nearly no loss of performance for large $\atr$ (green triangle). Finally, we find that the topic CVaR method we propose is close to the best possible \emph{oracle} performance.

\paragraph{Topic CVaR robustness beyond subpopulation shift.}
\begin{figure}[h]
  \centering
  \includegraphics[scale=0.5]{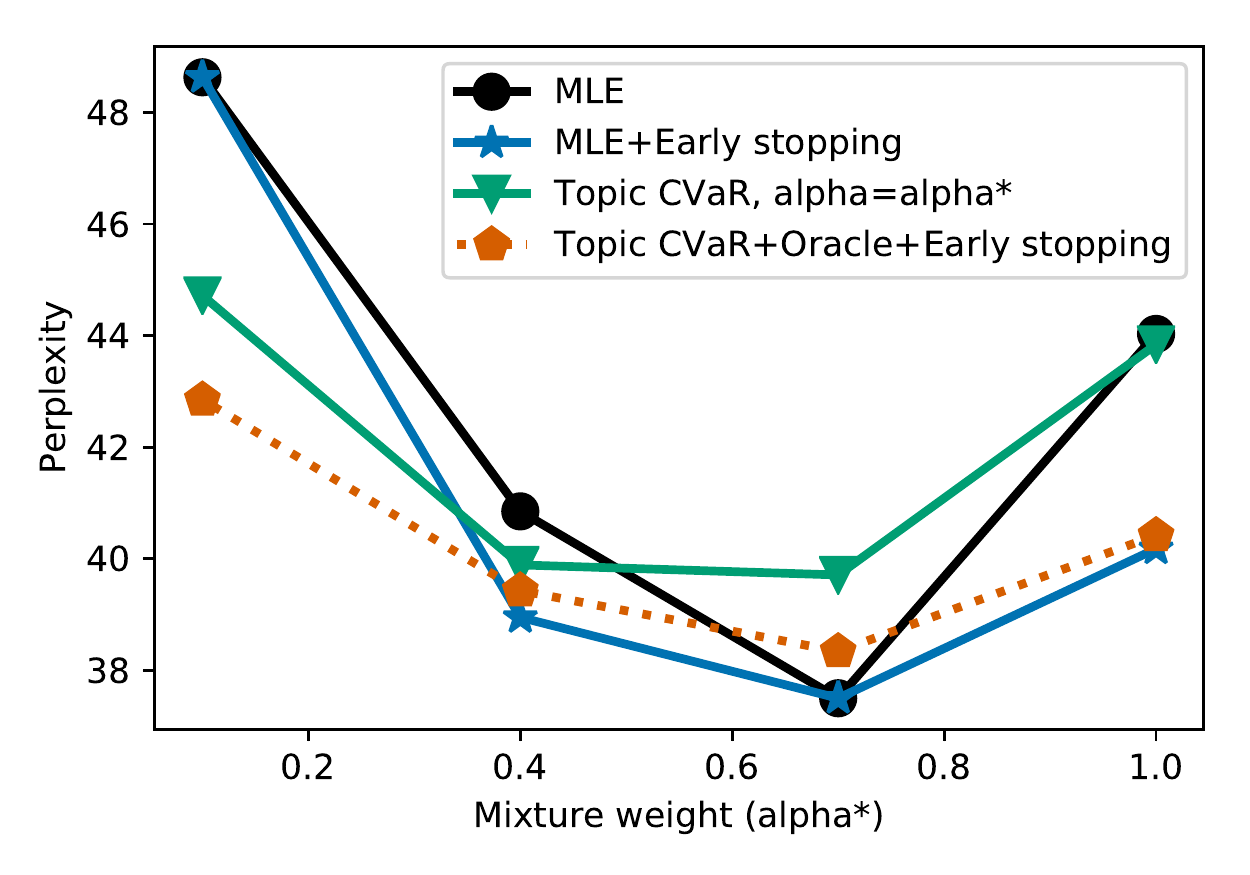}
  \caption{The robustness improvements from topic CVaR (black vs green and orange) apply even when the test set (\textsc{TripAdv} reviews) is not a subpopulation shift from the training set (\textsc{Yelp} and \lmdata{}). 
  }
  \label{fig:drotrip}
\end{figure}
The prior \textsc{Yelp}-\lmdata{} experiment showed that topic CVaR is more robust than MLE under subpopulation shift.

We now explore the more realistic setting in which the test distribution is not a subpopulation shift, but merely ``similar'' to the training distribution. 
We do this by testing the same model on the \textsc{TripAdv} hotel review corpus. The hotel and restaurant review distributions are similar (i.e. they both
frequently mention service) but differ in that hotels reviews often mention the
location and room, while restaurant reviews often mention food items.

We find a similar result consistent with the earlier subpopulation shift experiment (Figure \ref{fig:drotrip}). The MLE
model performance degrades rapidly between $\atr=0.7$ and $0.1$, while topic CVaR substantially reduces
this degradation. This suggests that topic CVaR models provide robustness benefits in real-world settings where the topic overlaps are not exact, and the subpopulation shift assumption no longer holds. 

\begin{figure}[h]
  \centering
  \includegraphics[scale=0.5]{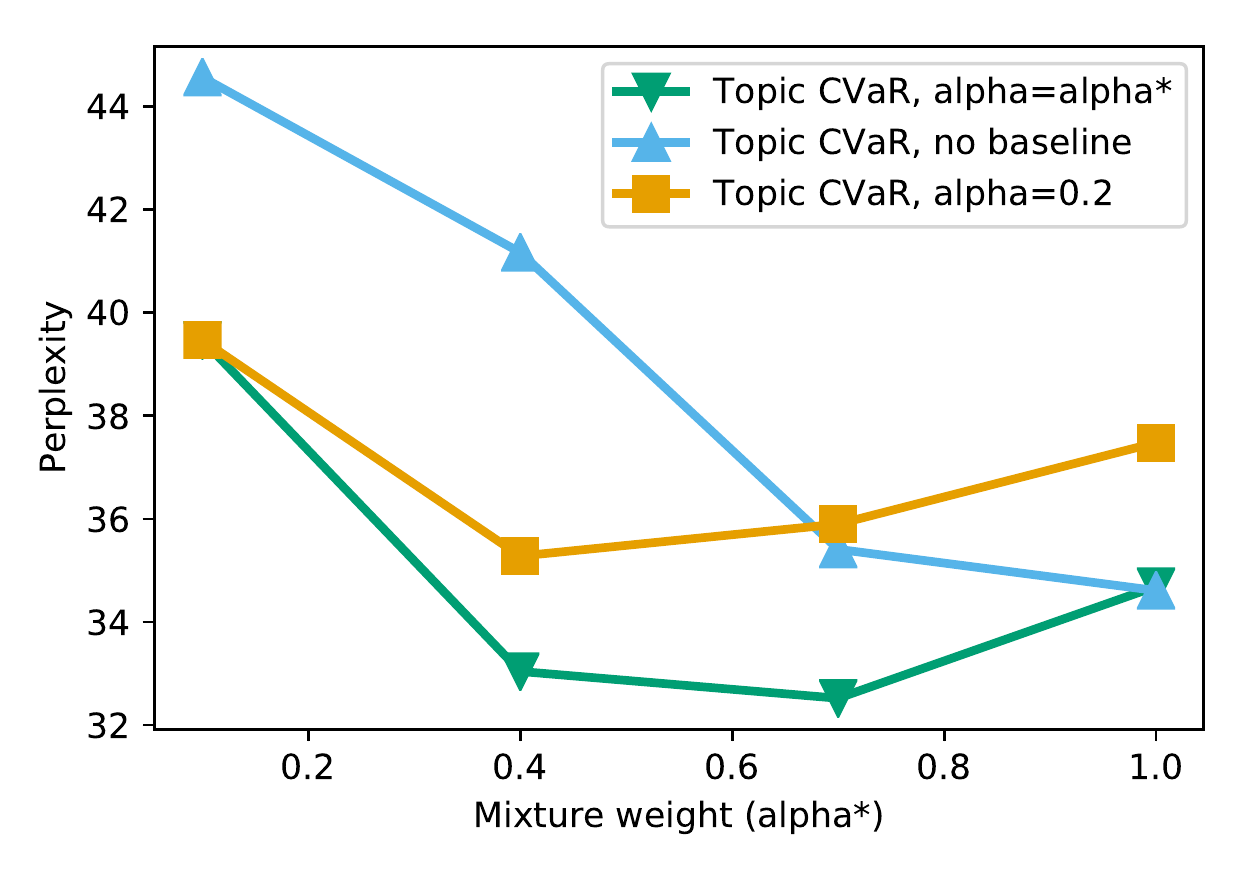}
  \caption{The robustness of topic CVaR degrades when the baseline is removed (teal), but is resistant to being over-conservative in choosing $\alpha$ (yellow).
  }
  \label{fig:droablation}
\end{figure}

\paragraph{Ablations.}
Topic CVaR extends the standard CVaR objective in two ways: the use of topics and the use of a baseline. We investigate the effect of these choices via an ablation experiment. Removing the topic structure results in dramatic loss of performance for our models: the perplexity exceeds 80 with $\alpha=0.2$ for all $\atr$. This is because the worst case group can consist of solely of disfluent sentences that do not match any real test distribution. If we remove the baseline, the resulting model is not completely degenerate, but it is not as robust as $\atr$ decreases (Figure \ref{fig:droablation}, teal). This is because \lmdata{} is a higher entropy corpus than \textsc{Yelp}, and forcing the model to achieve equal \emph{absolute} losses causes the model to focus nearly entirely on \lmdata{}, resulting in low \textsc{Yelp} performance.

\begin{figure}[h]
  \centering
  \includegraphics[scale=0.5]{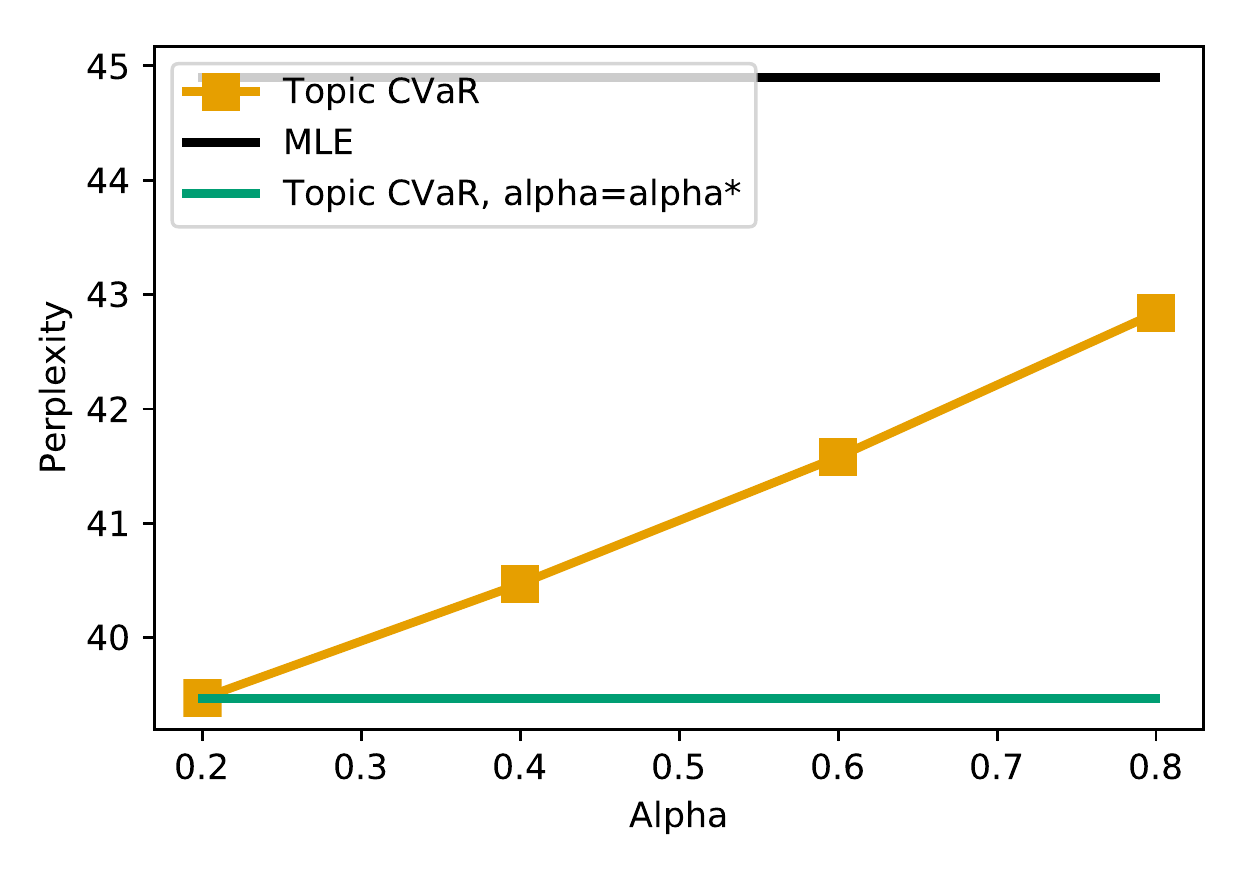}
  \caption{Topic CVaR outperforms MLE in the $\ptrainx \neq \ptestx$ setting ($\atr=0.1$) for any small $\alpha$ (x-axis). The performance degradation is linear, implying topic CVaR is robust to small errors in the choice of $\alpha$.
  }
  \label{fig:appl}
\end{figure}

\paragraph{Choice of $\alpha$.}
Since the true train-test overlap $\atr$ is not always known, we cannot always set our hyperparameter $\alpha$ equal to $\atr$. We find that selecting suboptimal values of $\alpha$ degrades perplexity between 2--3 points depending on $\atr$. Figure \ref{fig:droablation} shows that setting $\alpha$ to the most conservative choice of 0.2 outperforms MLE on small $\atr$ while incurring only 2 points of perplexity loss over MLE at $\atr=1.0$. Figure \ref{fig:appl} further demonstrates that when $\atr = 0.1$, any choice of $\alpha$ outperforms MLE, and incorrectly selecting $\alpha$ seems to incur a linear penalty in perplexity.

 \begin{table*}[ht]
   \centering 
     \resizebox{1.0\textwidth}{!}{
   \begin{tabularx}{1.4\linewidth}{p{0.65\textwidth} | p{0.65\textwidth}}
 
     \toprule
     $p_{\textbf{MLE}} > p_{\textbf{CVaR}}$ & $p_{\textbf{CVaR}} > p_{\textbf{MLE}}$ \\
     \midrule
     my girlfriend had an awful accident that hurt her leg \& ankle which resulted in a fire and rescue ride & huge servings, so plenty for leftovers.\\ \addlinespace[0.1cm]

     the address [PERSON] has listed is their old address & it tastes the way food should taste!\\ \addlinespace[0.1cm]

     wonderful location in a up and coming part of [GPE]. & every single person we spoke to on staff was absolutely incredible.\\ \addlinespace[0.1cm]

   \end{tabularx}
     }
     \caption{Examples from the \textsc{Yelp} corpus for which MLE outperforms topic CVaR (left column) and vice versa. Brackets indicate \textsc{OntoNotes} named-entity tags. The examples preferred by topic CVaR are stereotypical Yelp sentences, while those preferred by MLE refer to locations and accidents.
     }
     \label{tab:examplesacc}
     \vspace{-10pt}
 \end{table*}

\paragraph{Error analysis and examples.} 

\begin{figure}[h]
  \centering
  \includegraphics[scale=0.5]{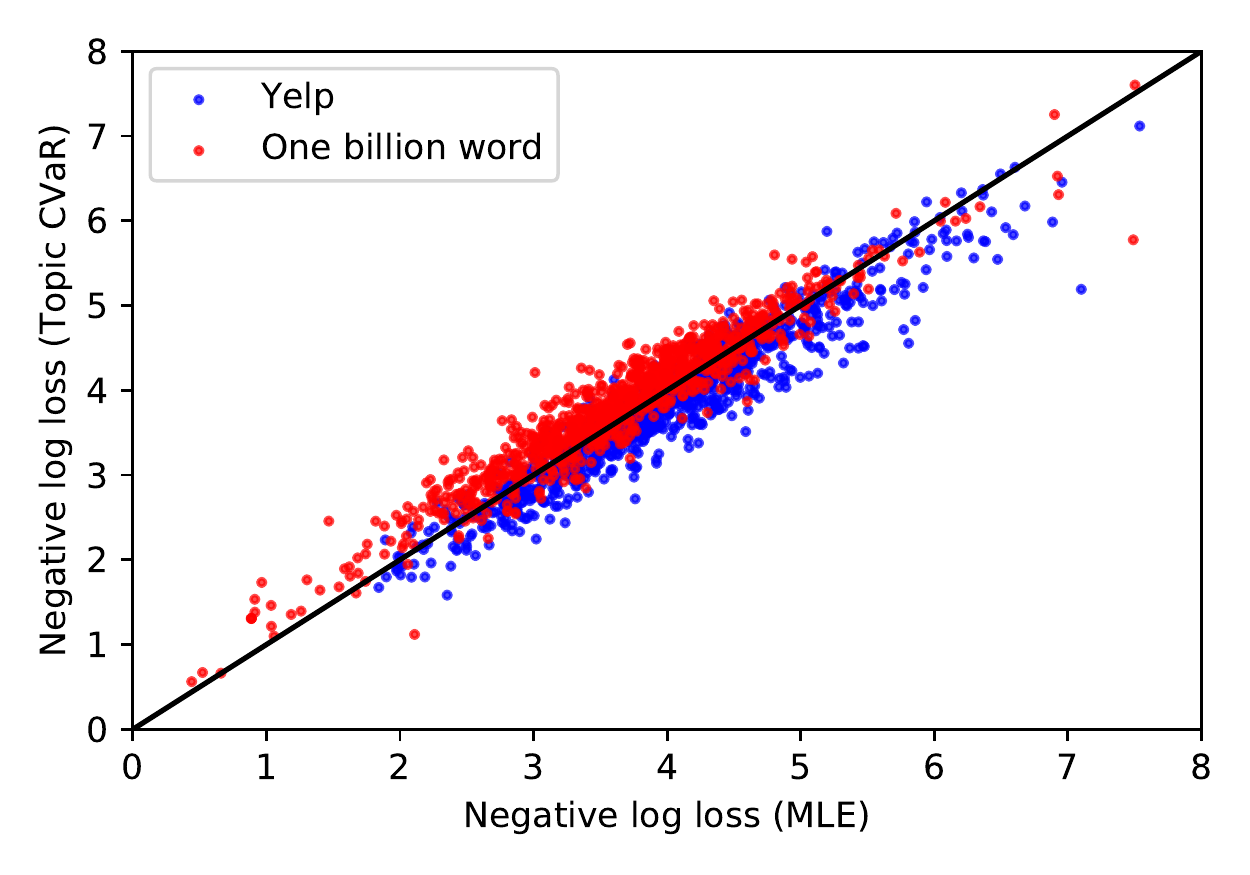}
  \caption{
    Log losses for sentences (points) from \textsc{Yelp} (blue) and \lmdata{} (red) under topic CVaR (y-axis) and MLE (x-axis). Topic CVaR performs well on \textsc{Yelp} and infrequent sentences (top right). MLE performs better on frequent sentences from \lmdata{} (bottom left). 
  }
  \vspace{-10pt}
  \label{fig:droscatter}
\end{figure}

Evaluating both models trained with $\atr=0.1$ on both the \textsc{Yelp} and \lmdata{} test sets, we find that topic CVaR assigns higher probabilities (and therefore incurs lower losses) on sentences from Yelp (Figure \ref{fig:droscatter}, top right). We also see that MLE does particularly well on low loss examples (bottom left) while topic CVaR does well on high-loss ones (top right) as we might expect from optimizing the worst-case losses. 

Examining examples from the \textsc{Yelp} test set (Table \ref{tab:examplesacc}), we identify examples which have substantially higher probabilities under MLE than topic CVaR (left column) and vice versa (right column). These examples show that topic CVaR performs well by assigning high probabilities to stereotypical \textsc{Yelp} sentences that discuss food and service, while MLE performs better on sentences about accidents and locations. These examples are consistent with the observation that topic CVaR assigns higher probabilities to typical \textsc{Yelp} sentences and thus has lower perplexity, while the MLE model has high perplexity since it assigns probabilities to \textsc{Yelp} sentences primarily based on their similarity to examples from \lmdata{}.

\section{Related Work}
\label{related}

\paragraph{Domain Adaptation:}
In the case of known source (train) and target (test) domains, there exist a variety of techniques to learn robust models \cite{shimodaira2000improving,quinonero2009dataset,daume07easyadapt,ben2010theory,blitzer2011domain, pryzant2017domainmix} or domain-invariant features \cite{ganin2015domain,tzeng2014domain}.
However, such methods require accurate domain membership annotations.

In the absence of domain membership annotations, prior multi-source domain adaptation \citep{mansour2009dams} approaches propose the use of clustering to identify candidate domains. For instance, \citet{hoffman2012discovering} and \citet{xiong2014latent} discover latent domains in classification by clustering data using class labels. \citet{gong2013reshaping} extend this work by identifying subsets which are distinct and learnable. More recent work consider
errors in estimating the target domain \citep{hoffman2018msda} and derive learning bounds with respect to such errors.
While these approaches make use of cluster and topic structures as prior, they still require some knowledge of the target distribution and train a model tailored to the target distribution.
Instead, we assume no knowledge on the target distribution and train a single model by considering the worst case.

In \emph{conditional} settings such as machine translation, prior works connect topic modeling and domain adaptation \cite{hu2014polylingual, eidelman2012topic}. However, unlike our work, these approaches use topics at \emph{test time} by inferring the domain from the input variable $x$. 
In language modeling, we have no inputs and thus must find models robust to unknown domain shifts at test time.
In addition, it can be difficult to infer the test distribution as the distribution can rapidly change across users and time.

\paragraph{Distributional Robustness:}
Our approach is based upon existing work in the distributionally robust optimization (DRO) literature. 
Optimizing on a ball of distributions around the empirical distribution has been considered in prior work \citep{bental2013robust, namkoong2017variance, duchi2016variance, sinha2018certifiable}.
Using DRO to minimize losses over subpopulations was proposed earlier in \citet{hashimoto2018repeated} and \citet{duchi2018learning},
and \citet{hu2018does} proposed incorporating problem structure via class labels. 
Our work derives an efficient optimization procedure for DRO with topic-based uncertainty sets, and demonstrates that naively applying DRO to log losses fails to provide robustness due to the lack of baselining. 

\section{Discussion}
\label{discussion}

In this work, we show that the performance of language models degrade as the amount of text from outside the test distribution grows. We hypothesize that this problem arises from the tendency of MLE to optimize for common sentences in the corpus, and we propose a solution based on distributionally robust optimization.

Empirically, we demonstrate that the DRO-based topic CVaR is more robust than MLE to subpopulation shifts and similar shifts.
While this work focuses on DRO for language modeling, train-test mismatches under subpopulation shifts are more broadly applicable to any task where there are trade-offs between potential test distributions, and potential test distributions can be described with topics. Our work shows that topics are an effective way to encode prior information about test distributions, and baselines can properly normalize for the difficulty across these topics.

\paragraph{Acknowledgments.} This work was supported by a PECASE Award and DARPA CwC program under ARO prime contract no. W911NF-15-1-0462. SS was supported by a Herbert Kunzel Stanford Graduate Fellowship. 
\paragraph{Reproducibility.} Code and data is available on CodaLab: \url{https://worksheets.codalab.org/worksheets/0xf8122ebd24e94209a2a1764007509098}.

\bibliography{refdb/all}
\bibliographystyle{acl_natbib}




\end{document}